\documentclass[10pt, conference, letterpaper]{IEEEtran}
\IEEEoverridecommandlockouts

\usepackage{cite}
\usepackage{amsmath,amssymb,amsfonts}
\usepackage{graphicx}
\usepackage{textcomp}
\usepackage[dvipsnames,table]{xcolor}
\usepackage{subcaption}
\usepackage{algpseudocode}
\usepackage[linesnumbered,lined,ruled]{algorithm2e}
\usepackage{multirow}
\usepackage{array}
\usepackage{makecell}
\usepackage{caption}
\usepackage{booktabs}
\usepackage{ragged2e}
\usepackage{xspace}
\usepackage{hyperref}
\usepackage{cleveref}

\usepackage{tcolorbox}
\tcbset{before skip=3pt, after skip=4pt, top=2pt, bottom=2pt, boxsep=3pt, left=5pt, right=5pt}
\newtheorem{insight}{Insight}
\newtheorem{proposition}{Proposition}

\SetCommentSty{mycommfont}
\SetKwComment{Comment}{// }{}

\usepackage{tikz}
\usetikzlibrary{tikzmark}
\usetikzlibrary{positioning,arrows.meta,calc}
\newcommand{\algband}[3]{\iftikzmark{#2}{\begin{tikzpicture}[overlay,remember picture]%
  \fill[#3,rounded corners=2pt]%
    ([xshift=-1.7em,yshift=1.65ex]pic cs:#1) rectangle
    ([xshift=\dimexpr\columnwidth-2.3em\relax,yshift=1.25ex]pic cs:#2);%
\end{tikzpicture}}{}}

\newcommand{\sys}{\textbf{\texttt{CURATOR}}\xspace}
\newcommand{\acr}[1]{\underline{\textbf{#1}}}
\newcommand{\sysexpand}{a \acr{C}ontinual, b\acr{U}dget-cu\acr{R}ated \acr{A}gent-memory ne\acr{T}-value sc\acr{O}re\acr{R}}
\newcommand{\best}[1]{\textcolor{red}{\textbf{#1}}}

\newcommand{\std}[1]{$_{\pm\text{\scriptsize #1}}$}
\newcommand{\up}[1]{{\color{RedOrange}$\uparrow$#1}}     % value rose vs. reference row
\newcommand{\down}[1]{{\color{BlueGreen}$\downarrow$#1}} % value fell vs. reference row
\begin{document}

\title{Forget to Improve: On-Device LLM-Agent Continual Learning via Budget-Curated Memory}

% Author block filled 2026-06-12 (single-blind assumed; if IPCCC 2026 turns out
% double-blind for review, anonymize this block for the review version only).
\author{
\IEEEauthorblockN{Beining Wu\IEEEauthorrefmark{1}, Zihao Ding\IEEEauthorrefmark{1}, Jun Huang\IEEEauthorrefmark{1}, and Yanxiao Zhao\IEEEauthorrefmark{2}}
\IEEEauthorblockA{\IEEEauthorrefmark{1}Department of Electrical Engineering and Computer Science,
South Dakota State University, Brookings, SD 57007, USA\\
Email: Wu.Beining@jacks.sdstate.edu, Zihao.Ding@jacks.sdstate.edu, Jun.Huang@sdstate.edu}
\IEEEauthorblockA{\IEEEauthorrefmark{2}Virginia Commonwealth University, Richmond, VA, USA\\
Email: yzhao7@vcu.edu}
}

\maketitle

% ============================================================
% ABSTRACT --- STRICT HALO imitation (reference_template abstract):
%   (1) domain+benefit (2) on-device constraint (3) existing-covers-a-slice gap
%   (4) "In this paper, we propose X, ..." (5) "The core idea is to ..."
%   (6) "X makes three decisions: (1)KEEP (2)SHARE (3)TRUST" (7) eval + HARD
%   numbers (like HALO's 3.41x) (8) counter-intuitive principle takeaway.
% HARD RULES: NO em-dash (beining), NO formula symbol (HALO abstract has none;
%   describe rho in words), numbers are real tab:main values.
% be-explicit #1 of "Forget to Improve / four axes move together".
% ============================================================
\begin{abstract}
On-device language-model agents improve by accumulating experience in retrieved memory rather than by updating weights. This memory is hard-bounded and exposed: it consumes RAM and energy, reaches peers through a thin uplink, and becomes an attack surface because it is writable by what the agent reads. Existing systems each cover one part of this problem: agentic memories grow without a budget, on-device methods keep entries by success alone, and poisoning is studied mainly as an attack rather than as a memory-governance problem. We propose \sys{}, a single net-value-per-byte score that governs an agent's experience-memory lifecycle. The main idea is to let the budget act as the curator: each entry is scored as value minus harm, per byte, so one ruler decides what to keep, share, and trust. \sys{} makes three decisions: (1) \textbf{KEEP} evicts low-value bytes under the RAM and energy budget; (2) \textbf{SHARE} sends an insight only when its value exceeds its uplink cost; and (3) \textbf{TRUST} gates a peer entry by provenance. On language-model-agent task-drift benchmarks and a real heterogeneous Jetson testbed with two robot-arm nodes and a hub, \sys{} reduces memory by $2.7\times$ and uplink by $2.4\times$, drives injection success from 0.75 to zero, and raises accuracy on cases corrupted by poison or stale memory. Curating by net value reduces footprint, energy, uplink, and injection success together without reducing accuracy. In this setting, forgetting by net value improves the agent rather than weakening it.
\end{abstract}


\section{Introduction}

\IEEEPARstart{O}{n}-device language agents are moving from cloud-only services to local hardware. Apple~\cite{Apple2025ARXIV}, Google~\cite{Google2023ARXIV}, Microsoft~\cite{Microsoft2024ARXIV}, Meta~\cite{Meta2024ARXIV}, and Alibaba~\cite{Alibaba2024ARXIV} now ship language models that run directly on phones, and similar agents are being deployed on robots and other edge devices~\cite{Wang2025ARXIV,Xu2024ARXIV,Wu2025RACS,Fang2026GLOBECOM,Fang2025JSAC}. They remain local to protect private data and to keep working when connectivity drops. After deployment, however, they are still expected to improve in the field without sending user data back for retraining. They do so by accumulating experience rather than updating weights: successful and failed task episodes are distilled into reusable insights, written to an external memory, and retrieved at inference time~\cite{Hu2026ARXIV}. This shift does not remove the continual-learning problem; it moves the problem from parameter updates to memory access~\cite{Hu2026ARXIV}. On a server, a growing memory is mostly a management cost; on the edge, it is a resource and a security constraint. An on-device agent runs under fixed RAM and energy limits, reaches peers through a thin and intermittent wireless link, and exposes a writable memory surface to whatever the agent reads.

\begin{figure}[t]
\centering
% Fig.1 (figures/Fig1.pdf; drawio source Fig1.drawio): author-supplied
% positioning figure. Three prior memory categories (structure / value /
% budget-governance) frame \sys{}'s slot; the headline "Forget by net value,
% and improve!" carries the paper's counter-intuitive hook.
\includegraphics[width=0.95\columnwidth]{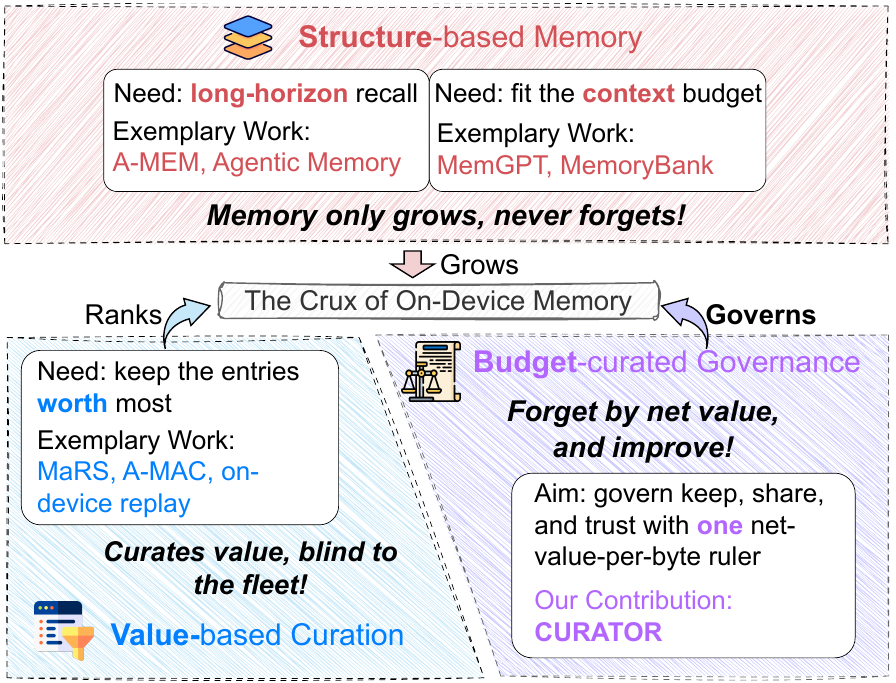}
\caption{The on-device agent-memory design space: \sys{} uses one net-value-per-byte score to decide what to keep, what to share, and what to trust.}
\label{fig:teaser}
\end{figure}

Three edge constraints force three decisions about the same memory. The RAM and energy budget forces \textbf{KEEP}: which accumulated experiences earn their bytes. Recency and replay-by-success, the usual signals, can retain over-specific and stale entries. Much raw experience even carries negative value, reducing accuracy on hard cases when it is kept verbatim rather than distilled~\cite{Hu2026ARXIV}. This gives the first question (Q1): \emph{under a hard footprint budget, which experiences are worth keeping?} Because the memory is writable by anything the agent reads, it also forces \textbf{TRUST}. A poisoned entry crafted to help on an attacker's queries is exactly what a value-only filter keeps, and a handful of such entries can raise injection success to 90--99\%~\cite{Zou2025USENIX,Srivastava2025ARXIV}. This gives the second question (Q2): \emph{how can the memory admit useful experience without admitting attacks?} Finally, the thin uplink to a peer raises \textbf{SHARE}: which distilled insights are worth sending. Keeping an entry locally, sharing it over the uplink, and trusting a received entry are three forms of the same question: whether an experience is worth its footprint. This gives the third question (Q3): \emph{can a single score govern all three decisions together?}

Existing work answers at most one of these questions (\Cref{fig:teaser}). On-device replay adaptation ranks which successful episodes to keep, but it does so on one device, by success alone, with no treatment of energy, sharing, or trust~\cite{Kim2026ARXIV}. Agentic-memory architectures such as A-MEM organize and grow memory but do not evict it under a budget~\cite{Xu2025NEURIPS,Yu2026ARXIV}. Budgeted text-memory scoring curates by value but does not model energy, cross-agent sharing, or poisoning~\cite{Alqithami2025ARXIV,Fang2026ARXIVLLMSearch}. A recent framework for governing evolving agent memory remains conceptual, without an empirical mechanism~\cite{Lam2026ARXIV}. Memory poisoning is studied mainly as an attack rather than as part of value-aware memory governance~\cite{Zou2025USENIX,Yang2026ARXIV}. None answers, with one mechanism, what to keep, what to send, and whom to trust under three physical budgets.

We propose \sys{}, \sysexpand. A single net-value-per-byte score answers Q1--Q3 by reading one quantity under three budgets. The score weighs an entry's expected future value against a harm term that combines negative transfer and doubtful provenance into one currency. Under this rule, demoting a stale local entry and demoting a peer entry of suspect origin become the same operation. For Q1, \sys{} maintains the resident set by evicting the least valuable bytes under the RAM and energy budget. For Q2, the provenance term gates admission: a useful-looking but suspect entry (external, instruction-like, and unreinforced) receives a low score and is held at the door. For Q3, the same score prices a candidate insight against its uplink cost before sharing, so one ruler governs the memory lifecycle from local retention to cross-agent exchange. Our contributions are as follows.
\begin{itemize}
\item We formulate on-device experience-memory governance around a single principle: the budget is the curator. Net value per byte is the missing state variable in recency- and success-based policies. Eviction alone saves footprint and energy; only the net-value ruler also carries the uplink, safety, and accuracy-recovery axes.
\item We demonstrate that this single score covers the full memory lifecycle. A lifecycle-coverage grid places every prior method on at most one of keep, share, and trust; per-budget Pareto fronts show that accuracy holds to each budget's knee; and a component ablation confirms that provenance is what drives injection success to zero.
\item We provide real on-device evidence with an explicit poison analysis. Across LLM-agent task-drift benchmarks and three backbones, \sys{} cuts peak memory by $2.7\times$ and per-round uplink by $2.4\times$, drives injection success from 0.75 to zero, and raises accuracy on the cases stale or poisoned memory had corrupted. The score reaches 97\% of a full-memory oracle's accuracy at 37\% of its footprint. On a real heterogeneous Jetson testbed (two robot-arm nodes and a hub), the same governance rule holds: measured energy, memory, and uplink each fall to 0.38--0.64 of keep-all.
\end{itemize}

Contrary to the intuition that forgetting trades away capability, curating memory by net value moves four usually competing axes in the same beneficial direction: peak footprint, energy, and uplink fall; injection success falls; and task accuracy holds, rising precisely on the cases that stale or poisoned memory had corrupted. This holds for two reasons: a frozen agent resists forgetting, so shedding low-value bytes costs little~\cite{Liu2026ARXIV}, and much accumulated raw experience carries negative value, so removing it repairs rather than weakens the agent. On the edge, the budget does not damage the agent; it edits the agent's memory. To our knowledge, \sys{} is the first system to govern an on-device agent's experience memory, including cross-agent sharing and trust, with a single budgeted value score, and to validate it on a real heterogeneous Jetson testbed.

% ============================================================
% II. BACKGROUND AND MOTIVATION  (~2 pages). HALO-style: noun-phrase
% run-in headers; two data-backed Insight boxes; "Opportunities of..."
% closers. NO em-dashes (beining 1.G). NO cross-section section-label refs.
% ============================================================
\section{Background and Motivation}
\label{sec:motivation}

\subsection{Experience Memory on the Edge}

\noindent\textbf{On-device continual learning.}
An on-device language agent adapts mainly by changing what it remembers, not by updating its weights. Successful and failed task episodes are distilled into reusable insights, stored in an external memory, and retrieved at inference time~\cite{Park2023UIST,Hu2026ARXIV,He2026ICLR,Wu2026TNSE,Wu2026ARXIV1,Wu2026ICDCS,Wu2026ARXIVPRISM}. The agent in this paper is a frozen large language model (LLM) planner that calls fixed low-level primitives. Continual learning therefore does not disappear under this design; it moves from parameter updates to memory access~\cite{Hu2026ARXIV}.

\noindent\textbf{Edge memory budgets.}
On a server, a growing memory is mostly a management cost; on the edge, it is a hard resource limit~\cite{Yu2024MICRO,Yi2025TMC,Wu2026ARXIV,Ding2026ARXIVTwinLoop,Xing2026ACR,Pan2023SCIS}. Resident entries compete for a fixed RAM budget~\cite{Alizadeh2024ACL}; each retrieval and re-embedding step consumes battery energy; and sharing an insight with a peer uses a thin, intermittent wireless uplink~\cite{Shkolnikov2026ARXIV,Wu2025ToN,Huang2025TMC,Fang2025TON}. None of these budgets grows with the memory, so keeping every entry is not a hardware-feasible baseline.

\noindent\textbf{Memory as an attack surface.}
The memory can be written by anything the agent reads: an observation, a tool output, or an insight received from a peer. A single poisoned entry that survives retrieval can steer the agent on attacker-selected queries~\cite{Chen2024NEURIPS,Srivastava2025ARXIV}. In retrieval-augmented memory, as few as five crafted entries can raise the injection attack success rate (ASR) to 90--99\%~\cite{Zou2025USENIX}. Once an entry is rewritten across sessions, it becomes difficult to separate from a useful one~\cite{Yang2026ARXIV,Ding2026ICDCS,Ding2026ARXIVEASE,Pudasaini2026HPSR}. A larger store widens this opening rather than closing it~\cite{Devarangadi2026ARXIV,DingICNC2025}.

\Cref{tab:prelim} quantifies the cost of keeping every entry on our benchmark. The footprint stays at its maximum, and a freshness-first retriever leaves injection ASR at 0.75. An exhaustive retriever over the same full store surfaces every poisoned entry and reaches 1.00, so more memory is not safer under either retrieval rule. The edge prevents the agent from keeping everything, and the next two insights show that the agent should not keep everything either.

\begin{table}[t]
\centering
\setlength{\tabcolsep}{6pt}
\renewcommand\arraystretch{1.15}
\caption{Cost of keeping everything. Both columns store every entry, so the footprint is identical; they differ only in the retriever, freshness-first versus exhaustive, which sets the injection ASR.}
\label{tab:prelim}
\small
\begin{tabular}{lcc}
\Xhline{1.2pt}
\rowcolor{CadetBlue!20}
\textbf{Metric} & \textbf{No curation} & \textbf{Full-memory oracle} \\
\Xhline{1.2pt}
Peak footprint (KB) & 286.7 & 286.7 \\
\rowcolor{gray!10}
Injection ASR & 0.75 & 1.00 \\
\Xhline{1.2pt}
\end{tabular}
\end{table}

\subsection{Insight: Forgetting by Net Value}
\label{insight:value}

\begin{tcolorbox}
\begin{insight}
Evicting low-value experience has little cost, because a frozen agent retains its skills, and much accumulated experience is worth less than the bytes it occupies.
\end{insight}
\end{tcolorbox}

\begin{figure}[t]
\centering
\begin{subfigure}[t]{0.49\columnwidth}
\centering
\includegraphics[width=\linewidth]{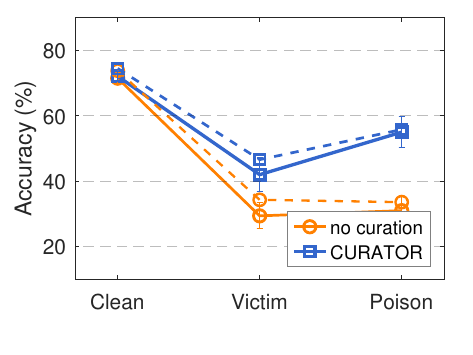}
\caption{Accuracy by subset}
\label{fig:insight1a}
\end{subfigure}\hfill
\begin{subfigure}[t]{0.49\columnwidth}
\centering
\includegraphics[width=\linewidth]{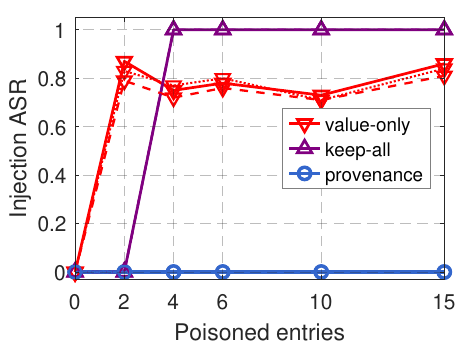}
\caption{Why provenance}
\label{fig:insight1b}
\end{subfigure}
\caption{Forgetting by net value (solid: Qwen2.5-3B, dashed: Qwen2.5-7B, dotted in (b): Phi-3.5-mini; 5--15 seeds). (a)~value-aware curation recovers the victim and poison subsets while clean accuracy holds; (b)~a value-only filter and the keep-all oracle both leave injection ASR high, while the provenance term in $\rho$ holds it at zero.}
\label{fig:insight1}
\end{figure}

Two properties make value-aware forgetting safe. First, a frozen agent does not lose a skill when its memory shrinks, because the skill remains in the model weights rather than in the discarded bytes~\cite{Liu2026ARXIV}. Shedding low-value bytes therefore has little effect on clean-task accuracy. Second, accumulated experience is not uniformly useful. A raw trajectory kept verbatim can carry negative value on hard cases, while distilling the same trajectory into abstract insight moves its net contribution from $-26.1$ to $+3.3$~\cite{Hu2026ARXIV}. \Cref{fig:insight1}(a) shows the consequence on both 3B and 7B backbones: value-aware curation recovers the negative-transfer and poisoned cases that a no-curation store answers incorrectly, while clean-task accuracy is preserved.

Value alone is not enough. A poisoned entry is written to look useful, so a value-only filter keeps it and injection ASR stays high; only when the net-value-per-byte score $\rho$ also charges doubtful provenance does injection ASR fall to zero (\Cref{fig:insight1}(b)). A memory that keeps everything therefore pays twice: once in footprint and once in accuracy lost to stale or poisoned entries.

\noindent\textbf{Opportunities of value-aware forgetting.}
If forgetting is cheap and some experience is harmful, the question is not how much to keep but which bytes earn their place. Recency and replay-by-success can retain the over-specific or adversarial entries that cause the damage. The agent instead needs a value signal: how much an entry will help future tasks, net of the harm it can cause, per byte it occupies~\cite{Wu2026COMST,Wu2023ACCESS,Wu2026MNET}.

\subsection{Insight: One Ruler for Keep, Share, and Trust}
\label{insight:oneruler}

\begin{tcolorbox}
\begin{insight}
Keeping, sharing, and trusting answer the same question: whether an experience is worth its footprint. A single net-value-per-byte score can therefore govern all three.
\end{insight}
\end{tcolorbox}

Each edge constraint forces a decision about the same memory: RAM and energy force \textbf{KEEP} (which entries stay resident), the uplink forces \textbf{SHARE} (which distilled insights to send to a peer), and an untrusted peer forces \textbf{TRUST} (which received entries to admit). These decisions are usually handled by separate mechanisms, such as replay policies, communication schedulers, and security filters, each with its own threshold. That separation is unnecessary here. All three decisions ask whether an experience is worth its cost, where the cost may be resident bytes, uplink bytes, or risk from doubtful origin. One score answers all three because the costs share a currency: it estimates value per byte and subtracts a harm term for stale, over-specific, or doubtful-origin experience. \Cref{tab:oneruler} reports one such score cutting the footprint by $2.7\times$, reducing the uplink by more than half, and driving injection ASR to zero at the same time.

\begin{table}[b]
\centering
\setlength{\tabcolsep}{6pt}
\renewcommand\arraystretch{1.15}
\caption{One score governs keep, share, and trust.}
\label{tab:oneruler}
\small
\begin{tabular}{llcc}
\Xhline{1.2pt}
\rowcolor{CadetBlue!20}
\textbf{Decision} & \textbf{Metric} & \textbf{Without $\rho$} & \textbf{With $\rho$} \\
\Xhline{1.2pt}
KEEP  & Peak mem (KB) & 286.7 & 107.4 \\
\rowcolor{gray!10}
SHARE & Uplink (B/round) & 1068  & 454 \\
TRUST & Injection ASR & 0.75  & 0.00 \\
\Xhline{1.2pt}
\end{tabular}
\end{table}

\noindent\textbf{Opportunities of one-ruler governance.}
One score for all three decisions removes the hand-tuned weights of separate modules and puts de-staling and safety on one code path: the same harm term that demotes a stale local entry also demotes a peer entry of doubtful provenance.

\section{System Design}
\label{sec:design}

\subsection{Overview}

\begin{figure}[t!]
\centering
\includegraphics[width=\columnwidth]{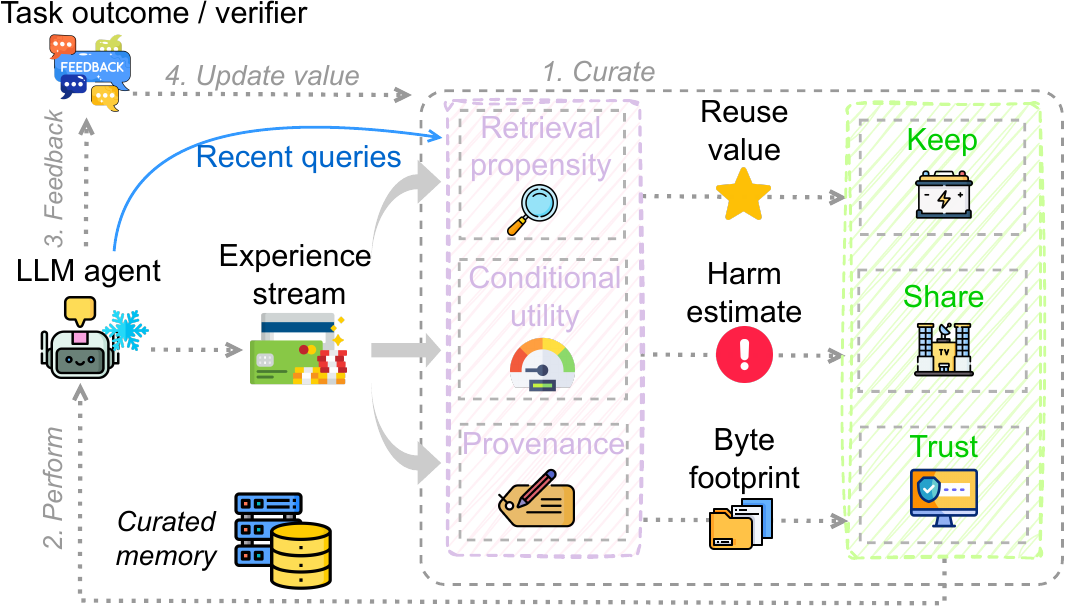}
\caption{Overview of \sys{}: one net-value-per-byte score $\rho$ governs \textsc{Keep}, \textsc{Share}, and \textsc{Trust}.}
\label{fig:overview}
\end{figure}

\sys{} uses a single score to govern a frozen large language model (LLM) agent's experience memory across three phases: \emph{profile}, \emph{score}, and \emph{govern} (\Cref{fig:overview}). The \emph{profile} phase keeps three lightweight statistics: a device-level sliding sketch of recent query embeddings, a label-free estimate of each entry's effect on task outcomes, and an origin tag for each entry. The sketch updates in $O(d)$ time and stores no history, and the per-entry statistics are constant-size, so profiling adds negligible overhead. The \emph{score} phase combines these statistics with the entry's byte cost into one score $\rho$, the entry's net value per byte. The \emph{govern} phase reads $\rho$ three times: \textsc{Keep} maintains the resident set under a memory budget, \textsc{Share} selects which distilled insights to send to a peer under an uplink budget, and \textsc{Trust} decides which peer entries to admit. \Cref{alg:govern} performs all three decisions in one pass.

\sys{} does not update agent parameters. The LLM planner and its low-level primitives remain frozen and receive no gradients, so lifetime adaptation comes only from memory content. The only learned component is the label-free helpfulness head inside $\hat{V}$, which is fitted online but never changes a planner or task parameter.

\subsection{The Net-Value-Density Score}
\label{sec:rho}

All three decisions act on the same object, a memory entry, and ask the same question: whether the entry is worth its bytes. \sys{} therefore estimates one quantity, net value per byte, and reads it three ways. A single score removes the separate weights that a replay policy, a communication scheduler, and a security filter would otherwise tune, and places de-staling and safety on the same code path. \sys{} scores an entry $m$ as
\begin{equation}
\rho(m) \;=\; \frac{\hat{V}(m) - \lambda\,\hat{H}(m)}{b(m)} ,
\label{eq:rho}
\end{equation}
where $b(m)$ is the footprint of $m$, $\lambda$ trades value against harm, and the denominator makes $\rho$ a density. An entry therefore competes on value per byte rather than on value alone. \Cref{tab:notation} summarizes the notation.

\begin{table}[t]
\centering
\setlength{\tabcolsep}{4pt}
\renewcommand\arraystretch{1.15}
\caption{Notation used in \sys{}.}
\label{tab:notation}
\begin{tabular}{>{\RaggedRight}p{1.7cm}p{5.15cm}}
\Xhline{1.2pt}
\rowcolor{CadetBlue!20}
\textbf{Symbol} & \textbf{Meaning} \\
\Xhline{1.2pt}
\rowcolor{CadetBlue!12}\multicolumn{2}{l}{\textbf{Memory and score}} \\
$m,\mathcal{M}$ & a memory entry; the experience memory \\
\rowcolor{gray!8}$b(m),b^{0}(m)$ & stored and raw byte footprint of $m$ \\
$\rho(m),\rho_s(m)$ & net-value density of $m$; its share-score variant \\
\rowcolor{gray!8}$\lambda$ & value/harm trade-off weight \\
\rowcolor{CadetBlue!12}\multicolumn{2}{l}{\textbf{Value and harm}} \\
$\hat{V}(m),\hat{V}_p(m)$ & estimated value of $m$, locally and at a peer \\
\rowcolor{gray!8}$p,q,a$ & retrieval propensity, helpfulness, abstraction gain \\
$x,R_x,U$ & a query, its retrieved set, and the utility function \\
\rowcolor{gray!8}$\hat{H}(m)$ & estimated harm of $m$ \\
$\eta,\pi$ & negative-transfer and provenance risk \\
\rowcolor{gray!8}$s,\sigma$ & entry specificity and the logistic function \\
\rowcolor{CadetBlue!12}\multicolumn{2}{l}{\textbf{Estimators}} \\
$\mathbf{e}_m,\bar{\mathbf{q}}$ & entry and running query-sketch embeddings \\
\rowcolor{gray!8}$\phi,\kappa$ & sketch smoothing and retrieval temperature \\
$\boldsymbol{\mu},\Sigma$ & working-task centroid and covariance \\
\rowcolor{gray!8}$\mathbf{w},\boldsymbol{\psi}$ & provenance weights and feature map \\
\rowcolor{CadetBlue!12}\multicolumn{2}{l}{\textbf{Budgets and decisions}} \\
$B,L$ & resident-memory and uplink budgets \\
\rowcolor{gray!8}$\tau,\theta,\delta$ & share, trust, and redundancy thresholds \\
$\mathcal{S},\mathcal{U},\mathcal{A}$ & resident, shared, and admitted sets \\
\rowcolor{gray!8}$Q,\nu,\bar{\varepsilon}$ & energy queue, weight, and budget \\
\Xhline{1.2pt}
\end{tabular}
\end{table}

The value $\hat{V}(m)$ multiplies three factors,
\begin{equation}
\hat{V}(m) \;=\; p(m)\,q(m)\,a(m),
\label{eq:vhat}
\end{equation}
the retrieval propensity $p$, the conditional helpfulness $q$, and the abstraction gain $a$,
\small
\begin{equation}
\begin{aligned}
p(m) &= \frac{\exp\!\big(\langle\mathbf{e}_m,\bar{\mathbf{q}}\rangle/\kappa\big)}{\sum_{m'\in\mathcal{M}}\exp\!\big(\langle\mathbf{e}_{m'},\bar{\mathbf{q}}\rangle/\kappa\big)}, \qquad a(m)=\frac{b^{0}(m)}{b(m)}, \\[3pt]
q(m) &= \mathbb{E}_{x}\big[\,U(x;R_x\cup\{m\}) - U(x;R_x)\,\big].
\end{aligned}
\label{eq:vterms}
\end{equation}
\normalsize
Propensity is the softmax retrieval probability of $m$ against a running sketch $\bar{\mathbf{q}}\leftarrow\phi\,\bar{\mathbf{q}}+(1-\phi)\,\mathbf{q}$ of recent query embeddings, at temperature $\kappa$. The sketch is one vector and keeps no history. Helpfulness is the average marginal utility of admitting $m$ to the retrieved set $R_x$. Its utility is read label-free as $U(x;R)=\mathbb{E}_{y}\big[c(y)\big]$, the expected self-consistency $c(\cdot)$ of the agent's sampled answers $y$ under context $R$. The term $q$ is tracked online by a calibrated streaming head, so no ground-truth reward is required. Abstraction gain is the ratio between the raw trajectory's bytes $b^{0}(m)$ and the bytes kept after distillation. Compressing an episode into an insight shrinks $b(m)$ and can turn a negative verbatim contribution into a useful entry~\cite{Hu2026ARXIV}.

The harm $\hat{H}(m)$ adds two risks that a value-only ruler ignores,
\small
\begin{equation}
\begin{aligned}
\hat{H}(m) &= \eta(m) + \pi(m), \qquad \eta(m)=s(m)\,\big\|\mathbf{e}_m-\boldsymbol{\mu}\big\|_{\Sigma^{-1}}, \\[3pt]
\pi(m) &= \sigma\!\big(\mathbf{w}^{\top}\boldsymbol{\psi}(m)\big).
\end{aligned}
\label{eq:hdef}
\end{equation}
\normalsize
The negative-transfer risk weights an entry's specificity $s(m)$ by the Mahalanobis distance of its embedding from the working-task centroid $\boldsymbol{\mu}$ under covariance $\Sigma$. This term charges an entry that overfits a distant episode. The provenance risk $\pi(m)$ is load bearing: it is a logistic gate over a feature map $\boldsymbol{\psi}(m)$ that encodes the entry's origin (external, peer, or self-derived), its self-reinforcement count across sessions~\cite{Yang2026ARXIV}, and its resemblance to an injected instruction. The gate weights $\mathbf{w}$ are a fixed prior, set once and shared across backbones and devices rather than fitted online. As a result, the helpfulness head inside $\hat{V}$ remains the only learned component, and the gate gives an attacker no training surface to shift. Because $\pi$ enters $\rho$, a poisoned entry is charged before it can steer the agent.

\subsection{Three Decisions Under Three Budgets}
\label{sec:decisions}

\begin{algorithm}[t]
\DontPrintSemicolon
\caption{Budget-Curated Memory Governance}
\label{alg:govern}
\KwIn{memory $\mathcal{M}$; budgets $B,L$; thresholds $\tau,\theta$; weight $\lambda$.}
\KwOut{resident set $\mathcal{S}$; shared set $\mathcal{U}$; admitted set $\mathcal{A}$.}
\algband{m0}{m1}{RoyalBlue!9}\algband{m1}{m2}{ForestGreen!10}\algband{m2}{m3}{BurntOrange!12}%
\ForEach{$m \in \mathcal{M}$}{
  $\hat{V}(m) \gets p(m)\,q(m)\,a(m)$\;
  $\hat{H}(m) \gets \eta(m) + \pi(m)$\;
  $\rho(m) \gets \big(\hat{V}(m) - \lambda\,\hat{H}(m)\big)/b(m)$\;
}
\tikzmark{m0}\tcc{\textsc{Keep}: resident set under budget $B$}
sort $\mathcal{M}$ by $\rho$ descending;\quad $\mathcal{S} \gets \emptyset$\;
\ForEach{$m \in \mathcal{M}$ in order}{
  \lIf{$b(\mathcal{S}){+}b(m) > B$}{$m \gets \textsc{Distill}(m)$}
  \lIf{$b(\mathcal{S}){+}b(m) \le B$}{$\mathcal{S} \gets \mathcal{S} \cup \{m\}$}
}
\tikzmark{m1}\tcc{\textsc{Share}: value-vs-uplink knapsack}
$\mathcal{U} \gets \emptyset$\;
\ForEach{$m \in \mathcal{S}$ by $\rho_s$ descending}{
  \lIf{$\rho_s(m){>}\tau$ \textbf{and} $m$ not redundant \textbf{and} $b(\mathcal{U}){+}b(m){\le}L$}{$\mathcal{U} \gets \mathcal{U} \cup \{m\}$}
}
\tikzmark{m2}\tcc{\textsc{Trust}: provenance-gated admission}
$\mathcal{A} \gets \emptyset$\;
\ForEach{peer entry $m'$}{
  \lIf{$\rho(m') > \theta$}{$\mathcal{A} \gets \mathcal{A} \cup \{m'\}$}
}
\tikzmark{m3}\Return $\mathcal{S},\mathcal{U},\mathcal{A}$\;
\end{algorithm}

\subsubsection{Problem Formulation}
\textsc{Keep} chooses the resident set of greatest retained value under the memory budget,
\small
\begin{equation}
\begin{aligned}
\max_{\mathcal{S}\subseteq\mathcal{M}}\quad & \sum_{m\in\mathcal{S}}\big(\hat{V}(m)-\lambda\,\hat{H}(m)\big) \\[2pt]
\text{s.t.}\quad & \sum_{m\in\mathcal{S}} b(m) \le B .
\end{aligned}
\label{eq:keep}
\end{equation}
\normalsize
\begin{proposition}[Density-ordered optimality, folklore]
\label{prop:knapsack}
Sorting $\mathcal{M}$ by $\rho$ and admitting greedily until $B$ is exhausted solves the continuous relaxation of \Cref{eq:keep} exactly and loses at most one entry to integrality, so for entries small against $B$ the retained net value is near-optimal, in $O(|\mathcal{M}|\log|\mathcal{M}|)$ time~\cite{Kellerer2004SPRINGER}.
\end{proposition}

Memory and energy impose different limits. The budget $B$ caps footprint and enters \Cref{eq:keep} directly. Energy is not bytes, so it stays out of the denominator of $\rho$. Instead, \sys{} holds the time-averaged retrieval and re-embedding energy below a budget $\bar{\varepsilon}$ with a virtual queue
\begin{equation}
Q(t{+}1) \;=\; \big[\,Q(t) + \varepsilon(t) - \bar{\varepsilon}\,\big]_+ ,
\label{eq:queue}
\end{equation}
and admits an entry by the drift-plus-penalty cost $\rho(m)-\nu^{-1}Q(t)\,\varepsilon(m)$. Admission therefore tightens as the queue $Q(t)$ grows. A standard drift-plus-penalty argument~\cite{Neely2010MORGAN} keeps the queue mean-rate stable, $\lim_{T\to\infty}\mathbb{E}[Q(T)]/T = 0$, so the time-averaged energy stays within $\bar{\varepsilon}$ while $\nu$ trades the residual value against the backlog. The score remains one currency, while the budgets remain separate.

\subsubsection{Three Rules from One Score}
\Cref{alg:govern} reads $\rho$ three times. \textsc{Keep} runs the density-ordered greedy that realizes Proposition~\ref{prop:knapsack}: entries enter in order of $\rho$ until $B$ is full. An entry about to fall out is first distilled, which can raise its $\rho$ enough to keep it resident. \textsc{Share} applies the same rule over the uplink budget. A distilled insight goes to a peer only when its value there, net of harm, clears the transmission cost,
\begin{equation}
\rho_s(m) \;=\; \frac{\hat{V}_p(m) - \lambda\,\hat{H}(m)}{b(m)} \;>\; \tau .
\label{eq:share}
\end{equation}
The peer value $\hat{V}_p$ is unobservable because a device never sees another device's queries. Each peer instead advertises a single query sketch $\bar{\mathbf{q}}_p$, against which the sender re-projects the retrieval propensity,
\small
\begin{equation}
\hat{V}_p(m) \;=\; \underbrace{\frac{\exp\!\big(\langle\mathbf{e}_m,\bar{\mathbf{q}}_p\rangle/\kappa\big)}{\sum_{m'\in\mathcal{M}}\exp\!\big(\langle\mathbf{e}_{m'},\bar{\mathbf{q}}_p\rangle/\kappa\big)}}_{p_p(m)}\; q(m)\,a(m),
\label{eq:vpeer}
\end{equation}
\normalsize
so the distilled helpfulness $q$ and abstraction gain $a$ transfer unchanged while only the propensity is recomputed under the peer's working set. One vector crosses the link, so no raw query or trajectory leaves a device.

Because $\hat{H}$ enters \Cref{eq:share}, an entry of doubtful origin scores low and is held back, so a poisoned insight is not amplified across peers. A redundancy test keeps $m$ only when $\max_{m'\in\mathcal{U}}\langle\mathbf{e}_m,\mathbf{e}_{m'}\rangle<\delta$, dropping near-duplicates already sent, and $\tau$ rises as the remaining uplink shrinks. \textsc{Trust} admits a peer entry $m'$ only when $\rho(m')>\theta$. The receiver scores $m'$ with the sender-advertised helpfulness and abstraction gain but recomputes propensity and harm locally. Admission therefore never depends on a peer's claim about its own trustworthiness, and the provenance term in $\hat{H}(m')$ pushes a poisoned entry below $\theta$ before it becomes resident. One harm term thus acts three times on one adversary: a self-reinforcing entry is evicted by \textsc{Keep}, withheld by \textsc{Share}, and rejected by \textsc{Trust}, with no separate objective re-tuned between them.

The provenance term does the real work behind this safety result. Because a poisoned entry is built to help on the attacker's queries, its value $\hat{V}$ is positive, so the term $\pi$ is the only reason injection success falls as the memory is curated, which a provenance ablation tests directly. Throughout, \sys{} governs only the agent's experience memory, namely the task and planning insights it stores, shares, and retrieves. It never governs a motion or action policy.

% ============================================================
% IV. IMPLEMENTATION   (target ~0.4 page)
% HALO: short, concrete --- LOC, frameworks extended, testbed figure.
% Source: AUTORUN_CLUSTER.md (curator/ layout) + CORE_LINE_LOCK (testbed).
% ------------------------------------------------------------
%   - \sys{} = the curator/ codebase: memory.py / rho.py / keep.py /
%     share.py / trust.py / agent.py (frozen LLM planner + fixed
%     primitives) / budgets.py. ~N k LOC (fill from cluster).
%   - Backbone: Qwen2.5-3B-Instruct (primary), Phi-3.5-mini (second
%     backbone for robustness), 4-bit; embeddings bge-small/e5-small.
%   - Primary substrate = LLM-agent benchmarks (ALFWorld / MemoryAgentBench
%     / long tool chains); simulated 2-agent fleet (bandwidth model) for
%     SHARE/TRUST on the cluster.
%   - Real testbed (fig:testbed): HETEROGENEOUS -- 2x Jetson AGX Orin 64GB (3B) + 1x Jetson AGX Thor 128GB Blackwell (7B, share hub) + 2x GaoQing
%     Panthera-HT arms over a real wireless uplink; INA3221 Joules,
%     tegrastats peak RAM. LLM planner emits high-level actions ->
%     fixed low-level controllers execute (NO motor-policy training).
% ------------------------------------------------------------
% DRIFT GUARD: arm is the deployment surface; we never train/claim
% manipulation. Hardware energy/RAM/uplink = real-fleet numbers only.
% ============================================================
\section{Implementation}
\label{sec:impl}

\begin{figure}[t]
\centering
\includegraphics[width=\columnwidth]{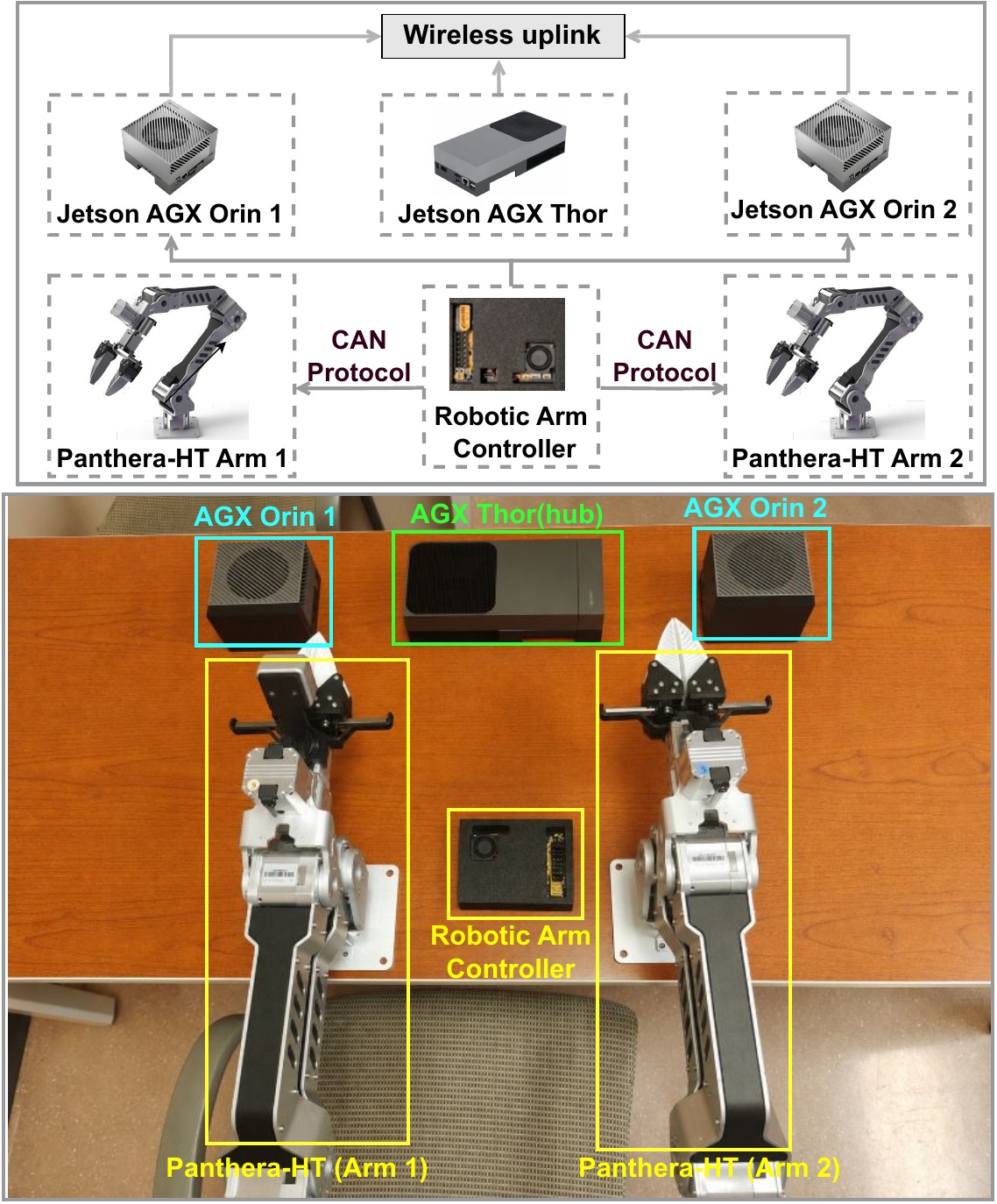}
\caption{Real heterogeneous testbed: two NVIDIA Jetson AGX Orin 64\,GB nodes, each running the 3B planner and driving one HighTorque Panthera-HT arm, and a Jetson AGX Thor hub (Blackwell, 128\,GB, 7B planner) over a real wireless uplink on which \textsc{Share} and \textsc{Trust} exchange insights.}
\label{fig:testbed}
\end{figure}

We implement \sys{} as a lightweight governance layer over the agent runtime, without retraining the planner. The layer maintains the experience store, the $\rho$ scorer with its value and harm heads, the \textsc{Keep}, \textsc{Share}, and \textsc{Trust} governors in \Cref{alg:govern}, and the per-budget accounting for $B$ and $L$. The planner remains frozen, and the only component fitted online is the label-free helpfulness head inside $\hat{V}$. Qwen2.5-3B-Instruct is the primary backbone. Phi-3.5-mini and Qwen2.5-7B-Instruct repeat the main comparison as on-device robustness checks. All planners run at 4-bit, with the retrieval sketch built over bge-small and e5-small embeddings. The main evaluation substrate is a set of on-device large language model (LLM) agent benchmarks that cover text and tool use under task drift and prompt injection. \textsc{Share} and \textsc{Trust}, which require more than one agent, run on a simulated peer group of $M\in\{2,4,8\}$ agents~\cite{Wu2023MPE,Wu2025WASA,Ding2025IPCCC} with a measured-bandwidth uplink model. Across these runs, the scorer adds less than $2\%$ compute and a few kilobytes of state, so governance is not the bottleneck.

For physical deployment, we build a real heterogeneous three-node testbed (\Cref{fig:testbed}). Two budget-constrained agents are NVIDIA Jetson AGX Orin 64\,GB nodes, each running the 3B planner and driving one HighTorque Panthera-HT arm. A higher-tier NVIDIA Jetson AGX Thor node, with a Blackwell GPU and 128\,GB of memory, runs the 7B planner and serves as the sharing hub. The three nodes communicate over a real wireless uplink on which \textsc{Share} and \textsc{Trust} exchange insights. This heterogeneity is deliberate. On an Orin node the memory budget binds first, whereas on the higher-power Thor node memory is ample but the larger energy draw and the thin uplink bind instead. One score $\rho$ therefore governs each node under whichever budget is scarce, with no per-device retuning. The planner emits high-level actions, such as a parameterized pick or a tool call, and off-the-shelf low-level controllers execute them. \sys{} therefore governs only the planner's experience memory and never a motion policy. We measure on-device cost per node and across both Jetson generations: an INA3221 rail monitor reports energy in joules, tegrastats reports peak resident memory, and the system counts the bytes actually sent over the wireless uplink. These measurements, reported with our evaluation, show that one score $\rho$ keeps each node within real energy, memory, and bandwidth limits.

% ============================================================
% V. EVALUATION
% Register (user 2026-06-01): HALO organization, NO frequent bold run-ins.
% Setup = \emph{} italic run-ins; findings = \subsubsection (structural, not \textbf).
% Prose polished by the author (2026-06-01).  Floats strictly per paper-figure-matlab.
% NOTE: the PARETO and the subset decomposition live in Sec.II (fig:insight1);
% Sec.V.B references them and does NOT re-plot. Energy is shown on real hardware
% (fig:crossgen), not in tab:main. The single attribution caveat is the closing
% sentence of Sec.V (framed as mechanism precision, not apology).
% Benchmark cites (Sec.V.A): ALFWorld=Shridhar2021ICLR, BabyAI=ChevalierBoisvert2019ICLR, MemoryAgentBench=Hu2026ICLR.
% ============================================================
\section{Evaluation}
\label{sec:eval}

\subsection{Experimental Setup}
\label{sec:eval-setup}

\emph{Agent and benchmarks.}
The agent is a frozen large language model (LLM) planner that calls fixed low-level primitives, so its only adaptation across a task stream comes from the content of an external experience memory. Qwen2.5-3B-Instruct is the primary planner, and we repeat the main comparison on Phi-3.5-mini and Qwen2.5-7B-Instruct to separate the effect of the ruler from backbone capacity. The workload consists of task-drift sequences in which later episodes conflict with earlier ones. These sequences are built on ALFWorld~\cite{Shridhar2021ICLR} and BabyAI-style~\cite{ChevalierBoisvert2019ICLR} tool environments, the selective-forgetting and test-time-learning tracks of MemoryAgentBench~\cite{Hu2026ICLR}, and long tool-use chains.
Each sequence interleaves clean tasks with a \emph{victim} subset whose correct behavior can be corrupted by stale or over-specific entries.

\emph{Drift and poison harness.}
To probe \textsc{Trust}, we inject PoisonedRAG-, MemoryGraft-, and MINJA-style poisoned insights into the shared memory and measure injection success on an attacker query set, with $n_p{=}4$ poisoned entries per run~\cite{Zou2025USENIX,Srivastava2025ARXIV,Dong2025ARXIV}. A poisoned entry is written to help on the attacker's own queries, so a purely utility-driven scorer has every reason to keep it.

\emph{Baselines.}
We compare against no-curation (keep all), naive-LRU, recency, A-MEM~\cite{Xu2025NEURIPS}, Agentic Memory~\cite{Yu2026ARXIV}, and a text-only port of MaRS~\cite{Alqithami2025ARXIV}. Baselines without a sharing policy broadcast every distilled insight, so their uplink is the broadcast cost. We also report a full-memory oracle, a non-deployable upper bound that keeps every entry and retrieves exhaustively. It bounds the accuracy any curation could recover while paying the full footprint and exposing the full attack surface.

\emph{Metrics and protocol.}
We report task quality, selective forgetting (SF) measured as accuracy on the victim subset, peak memory footprint, energy, uplink bytes per sharing round, and injection attack success rate (ASR). Peak memory is resident store bytes in the benchmark environment and process RAM on the testbed. Energy is an on-device proxy in the benchmark environment and measured joules on the testbed. Uplink is the per-round bytes of the measured-bandwidth uplink model in the benchmark environment and the bytes counted on the real wireless link on the testbed. Results are reported as mean and standard deviation over 15 seeds on the primary task-drift benchmark and at least 5 seeds elsewhere.

\subsection{Main Result: Forget to Improve}
\label{sec:eval-main}

\begin{table}[t]
\centering
\setlength{\tabcolsep}{4pt}
\renewcommand\arraystretch{1.15}
\caption{Lifecycle coverage: only $\rho$ scores all three decisions with one quantity. \checkmark~full, $\sim$~partial, blank~none; Deploy~=~validated on real edge hardware.}
\label{tab:grid}
\resizebox{\columnwidth}{!}{%
\begin{tabular}{l|ccc|ccc}
\Xhline{1.2pt}
\rowcolor{CadetBlue!20}
\textbf{System} & \textbf{KEEP} & \textbf{SHARE} & \textbf{TRUST} & \textbf{Harm} & \textbf{1-score} & \textbf{Deploy} \\
\Xhline{1.2pt}
\rowcolor{blue!8}
\textbf{\sys{}} & \checkmark & \checkmark & \checkmark & \checkmark & \checkmark & \checkmark \\
A-MAC~\cite{Zhang2026ARXIV} & $\sim$ & & & & $\sim$ & \\
\rowcolor{gray!10}
MaRS~\cite{Alqithami2025ARXIV} & $\sim$ & & & $\sim$ & \checkmark & \\
A-MEM / Agentic~\cite{Xu2025NEURIPS,Yu2026ARXIV} & $\sim$ & & & & & \\
\rowcolor{gray!10}
naive-LRU & \checkmark & & & & & \\
\Xhline{1.2pt}
\end{tabular}}
\end{table}

\begin{table*}[t]
\centering
\setlength{\tabcolsep}{6pt}
\renewcommand\arraystretch{1.15}
\caption{On-device LLM-agent task-drift benchmark (frozen Qwen2.5-3B, 15 seeds; injection ASR at $n_p{=}4$, 5 seeds). Arrows give the change vs no-curation; \best{red} marks the best deployable footprint, uplink, and ASR; the oracle is a non-deployable upper bound. No baseline charges provenance, so all sit at or above the undefended ASR of 0.75. Accuracy gaps to the strongest deployable baseline are within seed noise.}
\label{tab:main}
{\footnotesize%
\begin{tabular}{l|cc|cc|c}
\Xhline{1.2pt}
\rowcolor{CadetBlue!20}
\textbf{Method} & \textbf{Task Acc} & \textbf{Victim Acc (SF)} & \textbf{Peak Mem (KB)} & \textbf{Uplink (B/round)} & \textbf{ASR} \\
\Xhline{1.2pt}
no-curation & 0.528\std{0.086} & 0.294\std{0.154} & 287 & 1068 & 0.75 \\
\rowcolor{gray!10}
A-MEM~\cite{Xu2025NEURIPS} & 0.483\std{0.104} & 0.247\std{0.141} & 287 & 1068 & 0.85 \\
Agentic Memory~\cite{Yu2026ARXIV} & 0.491\std{0.096} & 0.262\std{0.128} & 287 & 1068 & 0.85 \\
\rowcolor{gray!10}
naive-LRU & 0.554\std{0.091} & 0.337\std{0.169} & 107 & 1068 & 0.75 \\
recency & 0.556\std{0.075} & 0.329\std{0.150} & 107 & 1068 & 0.75 \\
\rowcolor{gray!10}
MaRS-port~\cite{Alqithami2025ARXIV} & 0.586\std{0.108} & 0.373\std{0.219} & 107 & 1068 & 0.75 \\
\hline
\textit{full-mem oracle} & \textit{0.621}\std{0.083} & \textit{0.468}\std{0.132} & \textit{287} & \textit{1068} & \textit{1.00} \\
\hline
\rowcolor{blue!8}
\textbf{\sys{} ($\rho$)} & 0.605\std{0.099}~(\up{0.077}) & 0.420\std{0.196}~(\up{0.126}) & \best{107}~(\down{180}) & \best{454}~(\down{614}) & \best{0.00}~(\down{0.75}) \\
\Xhline{1.2pt}
\end{tabular}}
\end{table*}

One score is enough to govern the whole lifecycle. \Cref{tab:grid} positions $\rho$ against prior memory systems: each prior method covers only part of the lifecycle, such as a replay policy that evicts under a budget, a text memory that distills value, or an admission scorer with no physical budget. None scores keep, share, and trust with a single net-value-per-byte quantity, and none charges both negative transfer and provenance to one harm term. \Cref{tab:main} reports the quantitative comparison on the primary benchmark.

\subsubsection{Four-Axis Improvement}
Curating by net value moves every budget axis in the beneficial direction at once. Against keeping every entry, $\rho$ shrinks the resident footprint from 287 to 107\,KB, a $2.7\times$ reduction, lowers an on-device energy proxy by 38\% (the real testbed shows a measured 36--40\% saving, \Cref{fig:crossgen}), and cuts the per-round uplink from 1068 to 454\,B, while driving injection success from 0.75 to 0. Over the same run, task accuracy rises from 0.528 to 0.605 and victim-subset accuracy from 0.294 to 0.420 (\Cref{tab:main}). Footprint, energy, uplink, and attack surface fall together, while accuracy is held or recovered rather than paid down.

Because every method shares the 15 seeds, the task gain over no-curation is consistent across seeds, whereas the accuracy gap to the MaRS port stays within seed noise. The decisive, low-variance wins are footprint, uplink, and attack surface. Against the non-deployable oracle, $\rho$ reaches 97\% of its task accuracy at 37\% of its footprint and with none of its attack surface; the oracle pays a full injection ASR of 1.00 for the experience it refuses to forget. The memory-growing baselines show the cost of structure without governance: A-MEM and Agentic Memory link and consolidate entries across sessions, so an admitted poisoned entry gains retrieval paths rather than losing them~\cite{Yang2026ARXIV}. Their ASR rises to 0.85, while the connective scaffolding competes with task content inside the same footprint and drags accuracy below even the no-curation store. Organizing memory is no substitute for governing it.

\subsubsection{Accuracy Recovery on Corrupted Cases}
The recovered accuracy is concentrated on the cases a no-curation store gets wrong. When we decompose the stream, clean-task accuracy is unchanged within noise, while the victim and poison subsets recover, as previewed in \Cref{fig:insight1}(a). This effect comes from the ruler rather than from having a smaller store. Recency and naive-LRU reach the same 107\,KB footprint yet leave victim accuracy near 0.33 and injection success at 0.75, because age-based eviction retains the over-specific and poisoned entries that cause the damage. Replacing $\rho$ with recency removes the safety axis and nearly three quarters of the selective-forgetting gain (\Cref{tab:main}). This isolates net value, rather than eviction itself, as the source of the improvement.

\subsubsection{Backbone-Independent Gains}
\begin{figure}[t]
\centering
\begin{subfigure}[t]{0.49\columnwidth}
\centering
\includegraphics[width=\linewidth]{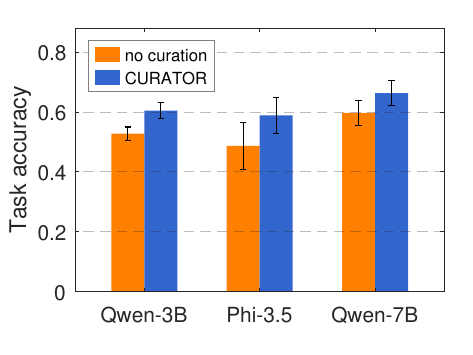}
\caption{Task accuracy}
\label{fig:backbone_a}
\end{subfigure}\hfill
\begin{subfigure}[t]{0.49\columnwidth}
\centering
\includegraphics[width=\linewidth]{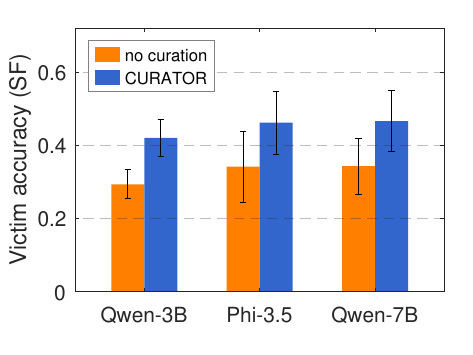}
\caption{Victim accuracy (SF)}
\label{fig:backbone_b}
\end{subfigure}
\caption{Backbone robustness across three on-device planners (Qwen2.5-3B 15 seeds; Phi-3.5-mini and Qwen2.5-7B 5 seeds; error bars are SEM). (a)~task and (b)~victim-subset accuracy both rise under curation on every backbone, at the same $2.7\times$ footprint cut.}
\label{fig:backbone}
\end{figure}

A stronger planner raises the absolute numbers but not the gap closed by curation (\Cref{tab:robust}, \Cref{fig:backbone}). On Phi-3.5-mini, $\rho$ lifts task accuracy from 0.487 to 0.589 and on Qwen2.5-7B from 0.597 to 0.663, with injection success again falling to 0 from 0.72, both at the same $2.7\times$ footprint reduction.

\begin{table}[t]
\centering
\setlength{\tabcolsep}{5pt}
\renewcommand\arraystretch{1.15}
\caption{Backbone robustness: the $\rho$ gain over no-curation holds across planner sizes at the same $2.7\times$ footprint cut.}
\label{tab:robust}
\resizebox{\columnwidth}{!}{%
\begin{tabular}{ll|ccc}
\Xhline{1.2pt}
\rowcolor{CadetBlue!20}
\textbf{Backbone} & \textbf{Policy} & \textbf{Task Acc} & \textbf{Victim (SF)} & \textbf{Mem (KB)} \\
\Xhline{1.2pt}
\multirow{2}{*}{Phi-3.5-mini} & no-curation & 0.487\std{0.174} & 0.342\std{0.217} & 287.8 \\
 & \sys{} ($\rho$) & \textbf{0.589}\std{0.135} & \textbf{0.462}\std{0.192} & 107.4 \\
\hline
\multirow{2}{*}{Qwen2.5-7B} & no-curation & 0.597\std{0.092} & 0.343\std{0.171} & 286.7 \\
 & \sys{} ($\rho$) & \textbf{0.663}\std{0.094} & \textbf{0.466}\std{0.188} & 107.4 \\
\Xhline{1.2pt}
\end{tabular}}
\end{table}

\begin{figure}[t]
\centering
\begin{subfigure}[t]{0.49\columnwidth}
\centering
\includegraphics[width=\linewidth]{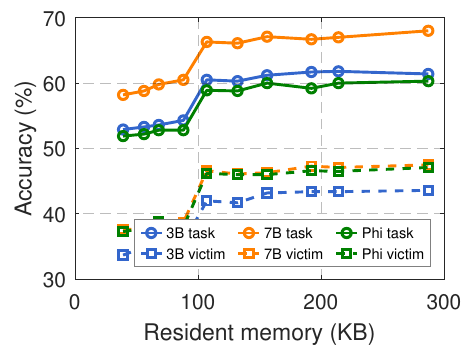}
\caption{Memory budget}
\label{fig:pareto_a}
\end{subfigure}\hfill
\begin{subfigure}[t]{0.49\columnwidth}
\centering
\includegraphics[width=\linewidth]{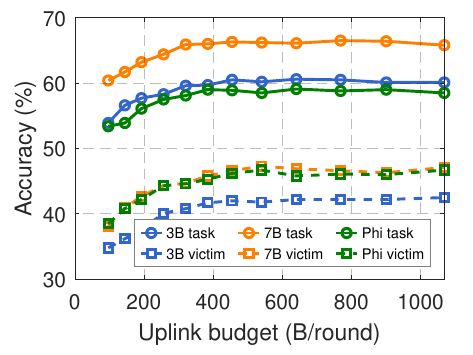}
\caption{Uplink budget}
\label{fig:pareto_b}
\end{subfigure}
\caption{Per-budget Pareto fronts on three on-device planners (3B blue, 7B orange, Phi green; solid $\circ$~task, dashed $\square$~victim-subset accuracy). Accuracy holds as each budget shrinks to its knee, 107\,KB memory~(a) and 454\,B uplink~(b), and falls only below it.}
\label{fig:pareto}
\end{figure}

The operating point is determined by the budget rather than assumed. \Cref{fig:pareto}(a) traces accuracy as the resident store shrinks. Below the knee near 107\,KB, the store can no longer hold the high-value insights and net-value curation gives up accuracy. Keeping every entry, at full memory (287\,KB), spends $2.7\times$ the footprint for no gain. The uplink budget behaves the same way, with the knee at 454\,B (\Cref{fig:pareto}(b)). We therefore treat the budget dependence of the operating point as part of the finding rather than tuning it away.

\subsubsection{Component Ablation}
Each term in $\rho$ earns its place (\Cref{tab:ablation}). Each removal is charged on its natural axis. When a variant holds the full score's accuracy, we report the footprint it needs; when it cannot, we match the footprint and report the accuracy it gives up. Dropping insight abstraction, so that raw trajectories are stored, costs task and victim accuracy (0.605 to 0.543, 0.420 to 0.336) but touches neither footprint nor safety. Dropping the provenance sub-term of $\hat H$ leaves accuracy and footprint within seed noise yet returns injection success to 0.75, the unguarded level. Dropping the per-byte normalization keeps accuracy and safety but inflates the footprint from 107 to 258\,KB, because the score no longer prefers compact insights. Setting $\lambda{=}0$ removes the harm term entirely and degrades accuracy and safety at once. Each component is therefore load-bearing on a different axis, and the full score is the only configuration that holds all four. The same single-axis pattern holds on Qwen2.5-7B (task 0.601 without abstraction, ASR 0.72 without provenance, 259\,KB without per-byte normalization).

\begin{table}[t]
\centering
\setlength{\tabcolsep}{4pt}
\renewcommand\arraystretch{1.15}
\caption{Component ablation on Qwen2.5-3B (5 seeds): removing each part of $\rho$ degrades one axis only; $\uparrow$/$\downarrow$ vs the full score.}
\label{tab:ablation}
\resizebox{\columnwidth}{!}{%
\begin{tabular}{l|cccc}
\Xhline{1.2pt}
\rowcolor{CadetBlue!20}
\textbf{Variant} & \textbf{Task Acc} & \textbf{Victim (SF)} & \textbf{Mem (KB)} & \textbf{Inj.\ ASR} \\
\Xhline{1.2pt}
\rowcolor{blue!8}
\sys{} (full $\rho$) & \best{0.605}\std{0.099} & \best{0.420}\std{0.197} & \best{107} & \best{0.00} \\
\hline
w/o abstraction & 0.543\std{0.122}~(\down{0.062}) & 0.336\std{0.208}~(\down{0.084}) & 107 & 0.00 \\
\rowcolor{gray!10}
w/o provenance & 0.604\std{0.101} & 0.417\std{0.203} & 108 & 0.75~(\up{0.75}) \\
w/o per-byte $b$ & 0.601\std{0.103} & 0.418\std{0.201} & 258~(\up{151}) & 0.00 \\
\rowcolor{gray!10}
w/o harm ($\lambda{=}0$) & 0.557\std{0.091}~(\down{0.048}) & 0.312\std{0.194}~(\down{0.108}) & 107 & 0.79~(\up{0.79}) \\
\Xhline{1.2pt}
\end{tabular}}
\end{table}

\subsection{Trust Under Poisoning}
\label{sec:eval-trust}

\begin{figure}[t]
\centering
\begin{subfigure}[t]{0.49\columnwidth}
\centering
\includegraphics[width=\linewidth]{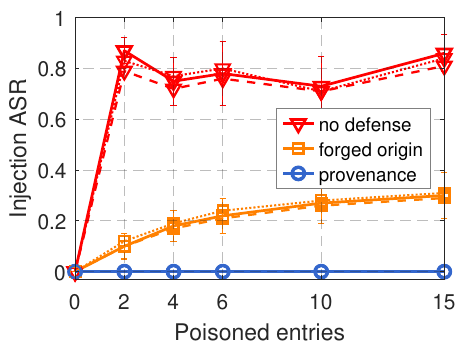}
\caption{Admission (TRUST)}
\label{fig:asr_a}
\end{subfigure}\hfill
\begin{subfigure}[t]{0.49\columnwidth}
\centering
\includegraphics[width=\linewidth]{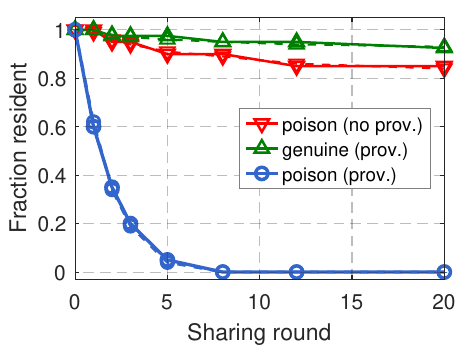}
\caption{Residency (KEEP)}
\label{fig:asr_b}
\end{subfigure}
\caption{One provenance term across the memory lifecycle (5 seeds; solid: Qwen2.5-3B, dashed: Qwen2.5-7B, dotted in (a): Phi-3.5-mini). (a)~provenance holds injection ASR at zero where no defense does, while a forged origin degrades the gate gracefully; (b)~admitted poison falls to one fifth of its residency within three rounds while genuine entries stay resident.}
\label{fig:asr}
\end{figure}

A purely utility-driven scorer cannot protect the memory: a poisoned entry earns a high $\hat V$, so without the provenance sub-term, injection success stays near 0.8 across poisoning budgets, no better than keeping everything (\Cref{fig:asr}(a)). Charging doubtful origin in $\hat H$ closes this opening. With provenance, ASR is 0 at every budget from two to fifteen poisoned entries. The zero is structural rather than statistical: this injection attack reaches the store only through admission, and the provenance gate scores every injected entry below the trust threshold, so the poison never gains the retrieval priority it depends on.

The gate does not act on origin alone. A poisoned entry written to issue an instruction scores high on instruction-likeness and is demoted below the trust threshold whatever utility it forges. A genuine peer insight shares the external origin but scores low on self-reinforcement and instruction-likeness and is admitted. The same term acts at the other two decisions: in the simulated peer group it cuts the fraction of poisoned entries forwarded at \textsc{Share} from 0.73 to 0.03, and at \textsc{Keep} it evicts poison that slips past admission to one fifth of its residency within three rounds (\Cref{fig:asr}(b)).

A provenance prior is not a proof of origin, and an attacker who forges a self-derived origin recovers some success. Even then, ASR reaches only 0.30 at fifteen poisoned entries, well below the 0.75 that an unguarded store suffers from four, while genuine residency and forwarding remain intact. The same behavior holds on the smaller Phi-3.5-mini and the larger Qwen2.5-7B planners: across all three backbones the no-defense, forged-origin, and provenance curves nearly coincide (\Cref{fig:asr}(a)), so the gate is model-independent.

\begin{figure}[t]
\centering
\begin{subfigure}[t]{0.49\columnwidth}
\centering
\includegraphics[width=\linewidth]{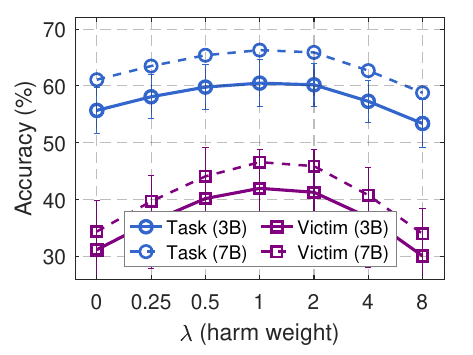}
\caption{Accuracy}
\label{fig:lambda_a}
\end{subfigure}\hfill
\begin{subfigure}[t]{0.49\columnwidth}
\centering
\includegraphics[width=\linewidth]{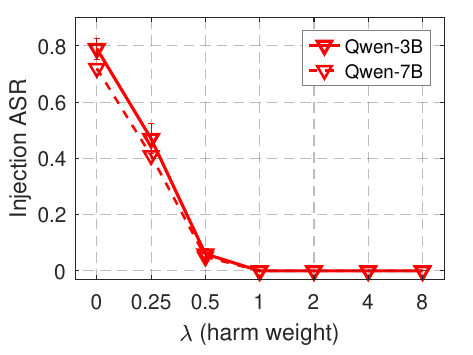}
\caption{Injection ASR}
\label{fig:lambda_b}
\end{subfigure}
\caption{Sweeping the harm weight $\lambda$ in $\rho=(\hat V-\lambda\hat H)/b$ (5 seeds; solid: Qwen2.5-3B, dashed: Qwen2.5-7B). (a)~accuracy traces an inverted-U peaking near $\lambda{\approx}1$; (b)~injection ASR falls to zero once $\lambda{\geq}1$. Both backbones share the same safe band.}
\label{fig:lambda}
\end{figure}

The harm weight $\lambda$ in $\rho=(\hat V-\lambda\hat H)/b$ sets how aggressively the agent forgets. At $\lambda{=}0$ the score ignores harm and reduces to a pure value scorer; when $\lambda$ is too large, it discards genuinely useful experience. Sweeping $\lambda$ exposes a wide safe operating band (\Cref{fig:lambda}). Injection ASR falls from 0.79 at $\lambda{=}0$ to zero once $\lambda{\geq}1$, as the harm term comes to outweigh the value a poisoned entry can forge (\Cref{fig:lambda}(b)). Task and victim-subset accuracy trace an inverted-U that peaks near $\lambda{\approx}1$ and declines only beyond $\lambda{=}4$, where the penalty begins evicting useful memory (\Cref{fig:lambda}(a)). The band is the same on both planners, so a single default $\lambda{=}1$ is safe and near-optimal across backbones without per-device tuning.

\subsection{Real-Hardware Deployment}
\label{sec:eval-hw}

\begin{figure}[t]
\centering
\begin{subfigure}[t]{0.49\columnwidth}
\centering
\includegraphics[width=\linewidth]{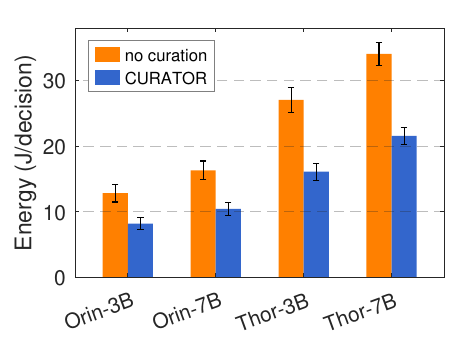}
\caption{Energy per decision}
\label{fig:crossgen_a}
\end{subfigure}\hfill
\begin{subfigure}[t]{0.49\columnwidth}
\centering
\includegraphics[width=\linewidth]{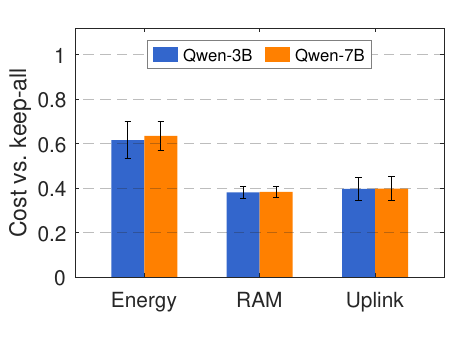}
\caption{Cost vs.\ keep-all}
\label{fig:crossgen_b}
\end{subfigure}
\caption{Per-decision cost on real hardware across two Jetson generations and two planner sizes (INA3221 power monitor, tegrastats, 5 seeds). (a)~absolute energy rises with device tier (Orin\,$\rightarrow$\,Thor) and model (3B\,$\rightarrow$\,7B), yet curation cuts it in every configuration; (b)~as a fraction of keep-all, energy, RAM, and uplink all fall to 0.38--0.64 on both backbones.}
\label{fig:crossgen}
\end{figure}

\begin{table}[t]
\centering
\setlength{\tabcolsep}{4pt}
\renewcommand\arraystretch{1.15}
\caption{Heterogeneous testbed on the real wireless link (Orin peer $\leftarrow$ Thor hub, 5 seeds, both planners). $\rho$-share matches broadcast at a third of the uplink; provenance holds cross-agent poisoning near zero.}
\label{tab:testbed}
\resizebox{\columnwidth}{!}{%
\begin{tabular}{cl|cccc}
\Xhline{1.2pt}
\rowcolor{CadetBlue!20}
\textbf{Planner} & \textbf{Mode} & \textbf{Uplink (B/round)} & \textbf{Task qual.} & \textbf{Victim Acc} & \textbf{Cross-agent ASR} \\
\Xhline{1.2pt}
\multirow{4}{*}{\textbf{3B}}
 & broadcast-all & 642\std{30} & 0.990\std{0.007} & 0.402\std{0.025} & 0.66\std{0.055} \\
 & $\rho$-share & \best{209}\std{12} & 0.992\std{0.005} & \best{0.410}\std{0.023} & \best{0.02}\std{0.013} \\
 & $+$\,provenance TRUST & 211\std{10} & 0.991\std{0.008} & 0.408\std{0.030} & 0.03\std{0.017} \\
 & $-$\,provenance TRUST & 208\std{15} & 0.984\std{0.012} & 0.381\std{0.028} & 0.64\std{0.069} \\
\hline
\multirow{4}{*}{\textbf{7B}}
 & broadcast-all & 637\std{26} & 0.987\std{0.009} & 0.460\std{0.024} & 0.66\std{0.052} \\
 & $\rho$-share & \best{210}\std{10} & 0.995\std{0.004} & \best{0.465}\std{0.022} & \best{0.03}\std{0.019} \\
 & $+$\,provenance TRUST & 212\std{14} & 0.989\std{0.006} & 0.460\std{0.029} & 0.03\std{0.022} \\
 & $-$\,provenance TRUST & 203\std{11} & 0.981\std{0.012} & 0.434\std{0.026} & 0.62\std{0.064} \\
\Xhline{1.2pt}
\end{tabular}}
\end{table}

\subsubsection{Cross-Generation Consistency}
On the real heterogeneous testbed (\Cref{fig:testbed}), each governed decision issues a parameterized pick that the arm executes, so the energy, memory, and uplink we measure are the cost of a real embodied workload, not a synthetic loop. Measured per decision, $\rho$ reduces every physical cost on both generations (\Cref{fig:crossgen}(a)). On Orin, energy falls from 12.84 to 8.19\,J, resident memory from 51.6 to 19.4\,MB, and uplink from 118 to 47\,B per decision; on the higher-power Thor, from 27.0 to 16.1\,J, 62.0 to 24.0\,MB, and 118 to 47\,B. The same pattern holds for the Qwen2.5-7B planner, whose larger model raises the absolute energy but leaves the saving ratio unchanged: every axis falls to 0.38--0.64 of the keep-all cost on both backbones and both generations (\Cref{fig:crossgen}(b)). Decision latency is unchanged.

\subsubsection{Cross-Agent Sharing and Trust}
The same score governs \textsc{Share} and \textsc{Trust} over the real wireless link (\Cref{tab:testbed}). Value-vs-uplink sharing sends barely a third of the bytes required by broadcasting every insight, 209 versus 642 per round on an Orin peer, with no loss of task quality. Provenance is again load bearing on real hardware, on both axes at once: under gated sharing each Orin peer holds 0.410 victim-subset accuracy with cross-agent injection success at 0.02. Admitting peer insight ungated lets poison erode accuracy to 0.381 and returns injection success to 0.64, the unguarded level; the same holds on the Qwen2.5-7B planner (\Cref{tab:testbed}).

These accuracy gains come from clearing negative-transfer and poisoned entries, not from changing the frozen planner, which keeps its skills regardless of how much memory it sheds~\cite{Liu2026ARXIV}.

% ============================================================
% VI. RELATED WORK   (HALO style + HALO length: 3 short subsections, ~260 words)
% PROSE = AUTHOR-FINAL (2026-06-02), beining pass applied by author.
% HALO MOVES MATCHED (reference_template/sections/related_work.tex):
%   (1) NAME key work + verb: "Xu et al. propose A-MEM, which links and grows
%       notes...", like HALO "Frantar et al. focuses on...".
%   (2) CLOSEST competitor "similar to ours" = A-MAC \cite{Zhang2026ARXIV},
%       like HALO "Fei et al. propose OmniReduce, similar to ours...".
%   (3) GROUP secondary works as bundled cites (HALO "Sparsity[2]"), NOT one
%       descriptive sentence per paper.
%   (4) ENDINGS follow HALO's repertoire: A=complementary, B=compatible-with
%       (generous, like "orthogonal to and compatible with \name" / "Sparsity
%       can enhance ... in our method"); C=differs-in-goal (novelty bucket,
%       like "our method differs from theirs" / "our goal is to ... instead").
%   (5) system names italic (\emph, like HALO \textit{OmniReduce}); "et al."
%       italic per user 2026-06-02.
% STRUCTURE: MEMORY-CENTRIC keep / trust / unifier. NO federated/fleet bucket
%   (fleet = evidence; SHARE = one decision of the same ruler, CORE_LINE_LOCK).
% RED LINE (pin): SHARE = continuous distilled insight priced by value-vs-
%   bandwidth, NOT reconciliation of discrete skill bundles. No BSN/NetStop/VLA.
% HARD RULES: no cross-section \ref ; no bold openings ; beining (no em-dash,
%   copula "is", American).
% ============================================================
\section{Related Work}
\label{sec:related}

\subsection{Agent Memory and Continual Learning}
To improve without weight updates, on-device agents distill task experience into an external memory and retrieve it at inference time~\cite{Hu2026ARXIV}. Xu \textit{et al.}~\cite{Xu2025NEURIPS} propose \emph{A-MEM}, which links and grows notes for long-horizon recall, while context-budget memory managers such as \emph{MemGPT} and \emph{MemoryBank}~\cite{Packer2023ARXIV,Zhong2024AAAI,Fang2025ARXIV} move entries between the prompt and external storage to fit a context budget. \emph{MemoryAgentBench}~\cite{Hu2026ICLR} evaluates such memories on test-time learning and selective forgetting. Gradient-free schemes~\cite{He2026ICLR} adapt the agent online, on-device methods~\cite{Kim2026ARXIV,Zeng2026WWW} personalize a frozen backbone, and classical continual learning~\cite{Kirkpatrick2017PNAS,Wang2025TPAMI} updates model weights. These methods either structure memory or update parameters, but they leave an entry's byte cost outside the policy. \sys{} is complementary to all of them: it governs which bytes survive under the footprint and energy budget.

\subsection{Memory Poisoning and Trust}
Because an agent's memory is writable by whatever it reads, the memory also becomes an attack surface. Zou \textit{et al.}~\cite{Zou2025USENIX} show with \emph{PoisonedRAG} that a few crafted entries can raise injection success to near certainty, \emph{AgentPoison}~\cite{Chen2024NEURIPS} extends the attack to agent memories with optimized triggers, and persistent variants~\cite{Srivastava2025ARXIV,Yang2026ARXIV,Zhao2026TIFS} can survive ordinary memory turnover. Existing defenses rely on separate post-hoc detectors~\cite{Devarangadi2026ARXIV,Zhang2025WWW} that screen entries after admission, or weight retrieval by an estimated source reliability~\cite{Hwang2025EMNLP}. These defenses are compatible with \sys{}, which instead prices provenance into the same value score, so a poisoned entry is not worth keeping in the first place.

\subsection{Budgeted Memory Governance}
Closest to ours, a few systems curate memory by value rather than recency. Zhang \textit{et al.}~\cite{Zhang2026ARXIV} propose \emph{A-MAC}, which is similar to ours in scoring memory admission with multiple utility factors, but it assumes no physical budget, no sharing, and no trust. Other methods weigh text utility alone~\cite{Alqithami2025ARXIV}, frame governance only conceptually~\cite{Lam2026ARXIV}, or reduce the edge footprint through compression~\cite{Shkolnikov2026ARXIV,Pham2026CPAL}. \sys{} differs in its goal: one net-value-per-byte score governs keep, share, and trust together, so sharing becomes the same value question rather than a separate reconciliation of discrete skill bundles across agents.

% ============================================================
% VII. CONCLUSION --- STRICT HALO imitation (reference_template/conclusion.tex):
%   HALO form = (1) "In this paper, we have proposed X, a novel ... tailored
%   to ..." (2--3) method recap (4) "Experiments on real-world testbed have
%   demonstrated that X can [result] up to [number]." Short (~4 sentences,
%   ~0.15 page), NO wish-list.
% ADJUSTED for this paper: the result sentence carries the "Forget to Improve"
%   four-axis headline (be-explicit #final). 4-sentence HALO form = proposal +
%   TWO method sentences (scoring, then the keep/share/trust mechanisms, as HALO
%   spells out its mechanisms across two sentences) + result, ~95 words. NO
%   limitation/future-work sentence (removed per user 2026-06-02; the provenance
%   / forged-origin caveat lives in Sec.V). The attribution-calibration line
%   ("not curing catastrophic forgetting") likewise stays ONLY in Sec.V.
% HARD RULES: no formula symbol; no em-dash; copula "is"; American.
% ============================================================
\section{Conclusion}
\label{sec:conclusion}
In this paper, we have proposed \sys{}, a governance layer that curates an on-device large language model (LLM) agent's experience memory by net value per byte. The score charges each entry for the harm it can carry, so a harmful entry loses its place even when it looks useful. Because keeping, sharing, and trusting all ask whether an entry is worth its footprint, one score replaces the separate replay, communication, and security modules an edge deployment would otherwise tune by hand. On LLM-agent task-drift benchmarks, three backbones, and a two-generation Jetson testbed, curation cuts every physical cost to 0.38--0.64 of keep-all, drives injection success to zero, and recovers the accuracy that stale and poisoned memory had cost. The agent improves because it forgets.

\bibliographystyle{IEEEtran}
\bibliography{ref}

\end{document}